\pdfoutput=1

\documentclass[11pt]{article}

\usepackage[]{acl}

\usepackage{times}
\usepackage{latexsym}

\usepackage[T1]{fontenc}

\usepackage[utf8]{inputenc}

\usepackage{microtype}

\usepackage{subfigure}
\usepackage{amsmath}
\usepackage{bm}
\usepackage{multirow}
\usepackage{graphicx}
\usepackage{enumerate}
\usepackage{breakurl}

\title{Other Roles Matter! Enhancing Role-Oriented Dialogue Summarization via Role Interactions}

\author{Haitao Lin\textsuperscript{1,2}, Junnan Zhu\textsuperscript{1,2}, Lu Xiang\textsuperscript{1,2}, Yu Zhou\textsuperscript{1,3}\thanks{\ \ Corresponding author.}, \\
{\bf Jiajun Zhang\textsuperscript{1,2}, Chengqing Zong\textsuperscript{1,2}} \\
\textsuperscript{1} National Laboratory of Pattern Recognition, Institute of Automation, CAS, Beijing, China \\
\textsuperscript{2} School of Artificial Intelligence, University of Chinese Academy of Sciences, Beijing, China \\
\textsuperscript{3} Fanyu AI Laboratory, Zhongke Fanyu Technology Co., Ltd, Beijing, China\\
\texttt{$\left\{\right.$haitao.lin, junnan.zhu, lu.xiang, yzhou, jjzhang, } \\
\texttt{cqzong$\left .\right\}$@nlpr.ia.ac.cn, }
}

\begin{document}

\maketitle

\begin{abstract}

Role-oriented dialogue summarization is to generate summaries for different roles in the dialogue, \textit{e.g.}, merchants and consumers. Existing methods handle this task by summarizing each role's content separately and thus are prone to ignore the information from other roles. However, we believe that other roles' content could benefit the quality of summaries, such as the omitted information mentioned by other roles. Therefore, we propose a novel role interaction enhanced method for role-oriented dialogue summarization. It adopts cross attention and decoder self-attention interactions to interactively acquire other roles' critical information. The cross attention interaction aims to select other roles' critical dialogue utterances, while the decoder self-attention interaction aims to obtain key information from other roles' summaries. Experimental results have shown that our proposed method significantly outperforms strong baselines on two public role-oriented dialogue summarization datasets. Extensive analyses have demonstrated that other roles' content could help generate summaries with more complete semantics and correct topic structures.\footnote{Our codes are available at: \url{https://github.com/xiaolinAndy/RODS}.}

\end{abstract}


\section{Introduction}
Dialogue summarization aims at compressing the main content of a long conversation into a short text. With the development of online conversation tools, the amount and length of conversation are growing up rapidly. Since a dialogue often contains complicated structure and ellipsis, it is time-consuming to read the whole dialogue. Dialogue summarization thus becomes valuable since it could extract the key point of a conversation and greatly reduce the time cost. This technique is widely used in customer service \citep{liu2019automatic}, meeting \citep{mccowan2005ami}, online chatting \citep{gliwa2019samsum}, etc.

\begin{figure}[tbp]
\centering
\includegraphics[width=\columnwidth]{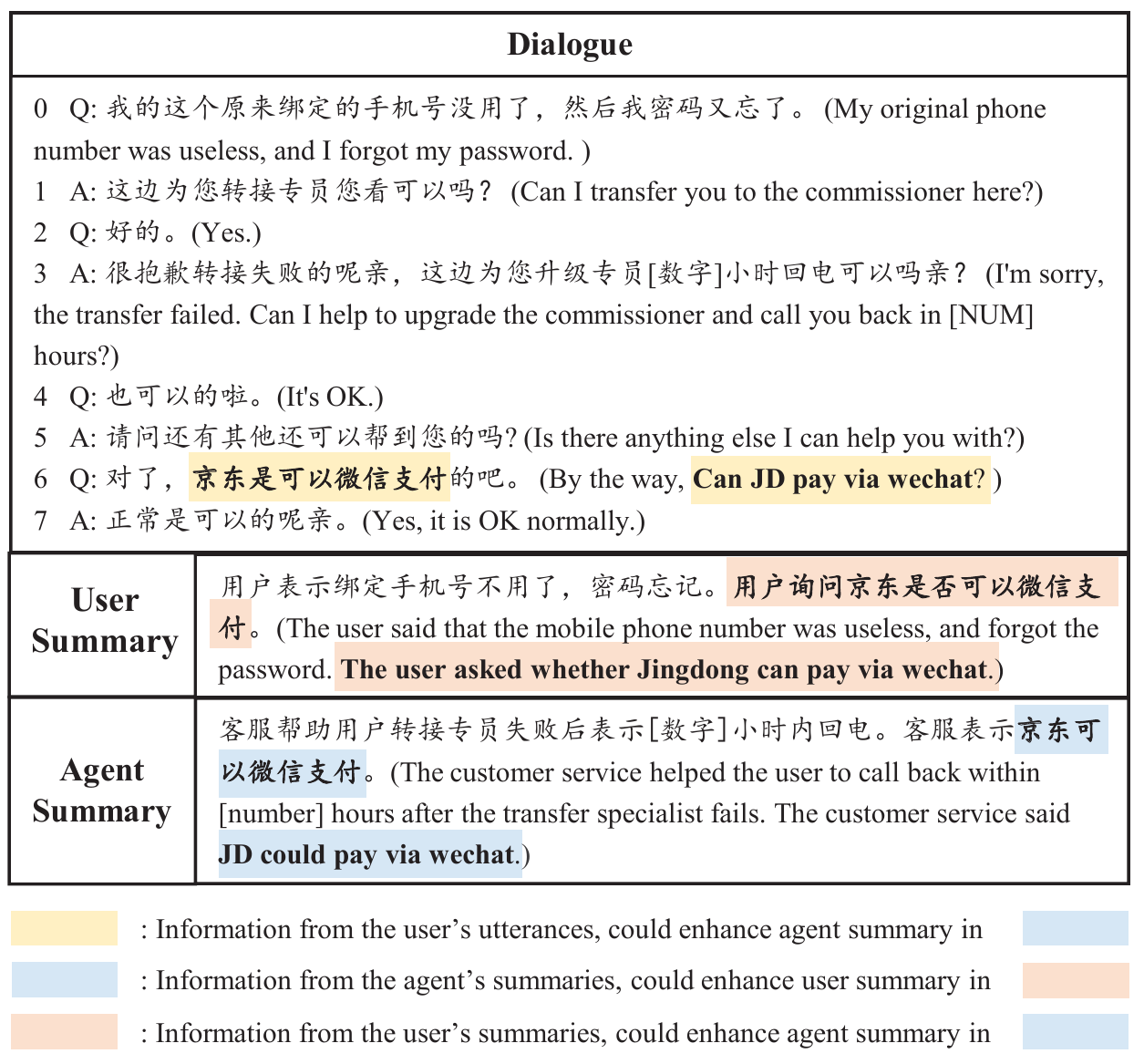} 
\caption{An illustration of the role-oriented dialogue summarization. The task will generate \emph{User Summary} and \emph{Agent Summary} for the user (Q) and the agent (A), respectively. The information from other roles could help enhance the summary quality.}
\label{fig:case1}
\end{figure}

In a dialogue, each role has its own opinion and goal, and different roles exchange information or reach a consensus through interactions. Therefore, in addition to summarizing the whole dialogue, we could summarize the main content for each role. \citet{tba} first define the role-oriented dialogue summarization task and provide a related dataset, CSDS. They define role-oriented dialogue summarization as grasping the main viewpoint of a given role from dialogue and mention the usage of role-oriented summaries in the customer service domain, \textit{e.g.}, reflecting the user's requirements and evaluating agent service quality. Besides, role-oriented summarization is beneficial to other dialogue domains such as medical inquiry \citep{song2020summarizing} and court debate \citep{10.1145/3357384.3357940}.

For role-oriented summarization, existing methods simply generate summaries for each role separately \citep{tba} or generate in a sequence labeling process \citep{song2020summarizing}. They ignore the strong relativeness among summaries for different roles and thus fail to utilize the information from other roles to enhance the summaries. However, information from other roles is also crucial for this task. We summarize two cases where other roles' information helps:

\textbf{(1) Other roles' dialogue utterances could help enhance the informativeness of summaries.} In Figure \ref{fig:case1}, utterance 7 (\emph{Yes, it is OK normally.}) is the key utterance of the agent's content, expressing a confirmation to the user's question. While only extracting it makes the agent summary ambiguous since it lacks the confirming object (\emph{JD can pay via wechat} in blue). In this case, the agent summary needs to integrate the content from the user's utterance (utterance 6 in yellow) to enhance its informativeness.

\textbf{(2) Other roles' summaries could help judge the key content in the dialogue.} In a dialogue, different roles often discuss the same topic. Therefore, considering the key content of the other role could help grasp the key content of a given role. As shown in Figure \ref{fig:case1}, the user summary contains a question about the payment (in red), and the agent summary contains the response to the payment question (in blue). If the summary of one role struggles in judging whether the discussion about payment should be contained in the summary, by referring to the summary of the other role, the summarization model could be more confident to include this information in the summary.

Although we notice the importance of other roles' information, it is difficult to extract the key information from other roles through a simple multi-task framework. The main issue is that it could not judge which information from other roles is useful without modeling the interaction between different roles. Thus, in this work, we propose two interaction methods to obtain key information from other roles for enhancing role-oriented summarization. First, we apply a cross attention interaction to let each role decoder select the most useful dialogue utterances from other roles. Specifically, we calculate the \textbf{Cross Attention} for different roles' utterances separately and add a new \textbf{Attention Divergence Loss} to interactively share the cross attention distributions between different roles. Second, we apply a decoder self-attention interaction to let each role decoder obtain other roles' summary information. We develop an interactive mechanism between decoders to consider other role summary information embedded in the decoder states. A new \textbf{Role Attention} module is added to each role decoder, where the attention object is the hidden states of other role decoders. At last, we use the role attention result and multiple context attention results to predict the word probability distribution of the summary. Through these two modules, the model could acquire more precise information from other roles and provide better role-oriented summaries.

To examine the effectiveness of our method, we conduct experiments on two dialogue summarization datasets \citep{tba,song2020summarizing} with role-oriented summaries in different domains (customer service, medical inquiry). We apply our method on two widely-used summarization frameworks (RNN-based and Transformer-based). The results have shown that, compared with baseline systems and naive multi-task approaches, applying role interactions could significantly improve the quality of role-oriented summaries. Further analyses verify that our proposed method can help the model correctly attend to other roles' key information and generate summaries with more complete semantic and correct topic structures.

The main contributions of this paper include: (1) We are the first to enhance role-oriented dialogue summarization by focusing on other roles' key information. (2) We innovatively design two role interaction methods to obtain other roles' key information useful for generating summaries. (3) Experimental results on two datasets have shown that our method could lead to considerable improvements. Besides, our method has good generalizability since it works on multiple baseline frameworks.

\section{Related Work}

\subsection{Dialogue Summarization}

Dialogue summarization has been studied in various domains, \textit{e.g.}, meeting \citep{mccowan2005ami,janin2003icsi}, daily chatting \citep{gliwa2019samsum,chen-etal-2021-dialogsum}, customer service \citep{liu2019automatic,Zou_Zhao_Kang_Lin_Peng_Jiang_Sun_Zhang_Huang_Liu_2021}, and medical inquiry \citep{song2020summarizing,krishna-etal-2021-generating}. Considering the particularity of dialogue, many studies try to improve the dialogue summarization performance by focusing on dialogue-specific features \citep{feng2021survey}, such as topic information \citep{chen-yang-2020-multi}, discourse structure \citep{chen-yang-2021-structure}, coreference information \citep{liu-etal-2021-coreference} and speaker information \citep{9414547, zhu-etal-2020-hierarchical}.

However, all the above studies focus on summarizing the whole dialogue. Only a few studies pay attention to role-oriented summarization, which aims to summarize the main content of a single role in the dialogue. A relative task is focused meeting summarization \citep{wang-cardie-2013-domain, mehdad-etal-2014-abstractive, zhong-etal-2021-qmsum}. It aims to summarize a specific part of the meeting dialogue, while role-oriented summarization focuses on a single role, and the relationship between different roles is much closer. \citet{tamura-etal-2011-extractive} focus on contact center dialogue summarization, but they only extract salient sentences from the dialogue and do not summarize for different roles. 

Due to the lack of labeled data, \citet{Zhang_Zhang_Zaheer_Ahmed_2021} propose an unsupervised method to generate summaries for the customer and the agent under a variational auto-encoder framework. As for supervised methods, there are only two datasets available for training. \citet{tba} propose a customer service domain dataset named CSDS, where each dialogue has an overall summary and two role-oriented summaries for user and agent. They train two separate models for generating user summaries and agent summaries. \citet{song2020summarizing} provide a medical inquiry dialogue summarization dataset where each dialogue has two extractive summaries for the patient and the doctor. They train a sequence labeling model to extract summaries for these two roles. Compared with these approaches, to the best of our knowledge, we are the first to enhance role-oriented summarization by explicitly considering other roles' critical information.

\subsection{Interactive Decoding}

Interactive decoding is a mechanism to share information between different decoders in the decoding process. \citet{10.1162/tacl_a_00256} propose this mechanism and use it on machine translation to simultaneously decode from both left-to-right and right-to-left. \citet{wang-etal-2019-synchronously} and \citet{Liu_Zhang_Xiong_Zhou_He_Wu_Wang_Zong_2020} further utilize it on more complex machine translation tasks, including multilingual translation and speech translation. In this work, we first use the interactive decoding mechanism on the summarization task to decode summaries for different roles, aiming at utilizing other roles' summary information for summarization. Besides, we also propose an interaction method on cross attention to utilize other roles' critical dialogue utterance information.

\section{Our Approach}

\subsection{Task Definition}

Given a dialogue $D$ containing $m$ utterances $\{u_1, ..., u_m\}$ and $p$ speakers $S = \{s_1, ..., s_p\}$, the role-oriented summarization task aims to generate a summary $y^k$ for each speaker $s_k$. Each utterance $u_k$ consists of a speaker role $r_k \in S $ and related content. By concatenating all the utterances and related speaker roles, we achieve the final input $\{x_1, ..., x_n\}$. Note that since both datasets used in this work have two speakers, one asking questions and one giving answers, we thus use $y^{user}$ and $y^{agent}$ to represent two role-oriented summaries in the following illustration\footnote{Here we need to point out that our method could also apply to dialogues with more than two speakers.}. 

\begin{figure*}[tbp]
\centering
\includegraphics[width=\textwidth]{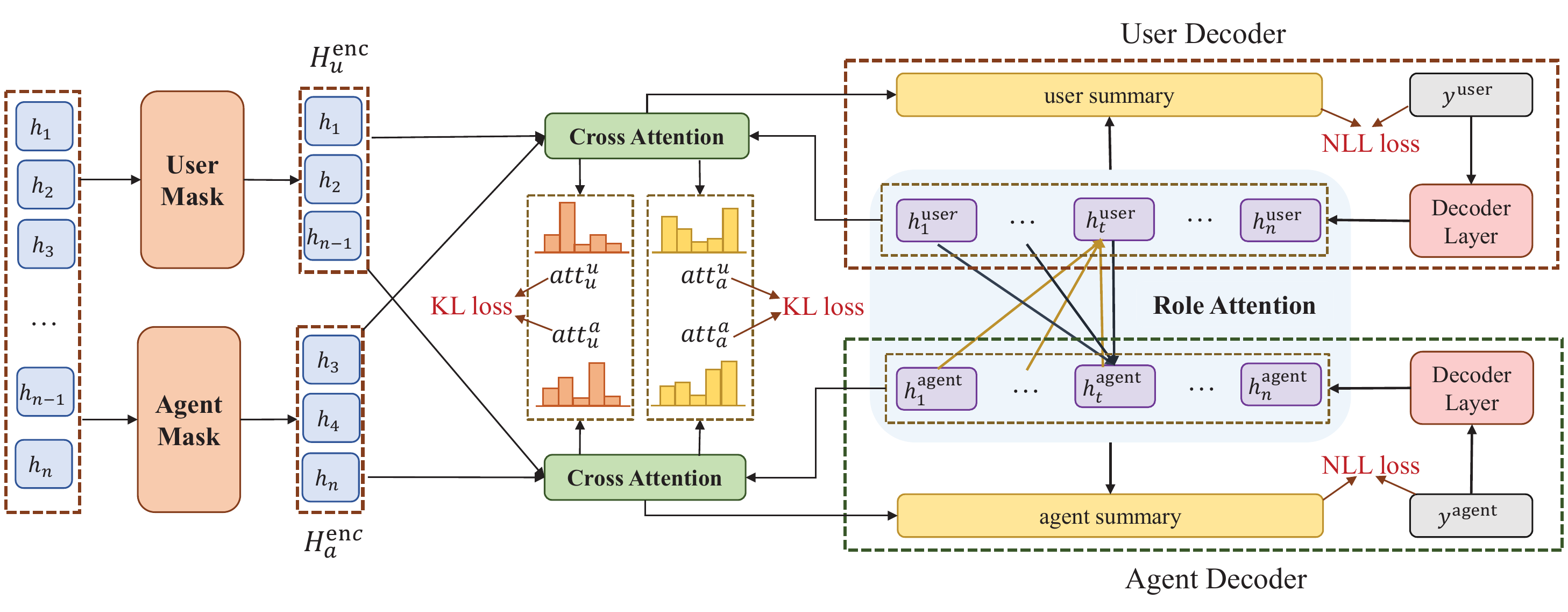} 
\caption{The model structure of our proposed method with role interactions.}
\label{fig:model}
\end{figure*}

\subsection{Role Interactions}

In a traditional encoder-decoder framework for dialogue summarization, the encoder hidden states represent the semantic information of input dialogue utterances, and the decoder hidden states contain the information used to generate summaries. To fully exploit the information from other roles, we apply two role interactions on the attention module of both hidden states. We present the structure of our method in Figure \ref{fig:model} and introduce the details of interactions in the following paragraphs.

\subsubsection{Cross Attention Interaction}
Our method is constructed based on a multi-task framework where an encoder is used to encode dialogue utterances and two role decoders (user decoder and agent decoder) are used to decode user summary and agent summary. First, the input $\{x_1, ..., x_n\}$ is sent to an encoder (omitted in the figure for simplicity) and the encoder outputs the context hidden representation $\{h_1, ..., h_n\}$. In the decoding phase, to calculate the cross attention results for different roles separately, we use \textbf{User Mask} and \textbf{Agent Mask} to split the context information into user context $H^{\mathrm{enc}}_u$ and agent context $H^{\mathrm{enc}}_a$. $H^{\mathrm{enc}}_u$ contains the hidden representation of all user utterances, and $H^{\mathrm{enc}}_a$ contains the hidden representation of all agent utterances.

The cross attention module extracts the most useful information from the context according to the temporary decoder state. Here we modify the module to attend to different role contexts separately. Taking user decoder as example, at step $k$, we use the hidden state of user decoder $h^{\mathrm{user}}_k$ to attend to user context $H^{\mathrm{enc}}_u$ and agent context $H^{\mathrm{enc}}_a$, obtaining two attention distributions $att_{u, k}^u$, $att_{a, k}^u$ and context attention results $c_{u, k}^u$, $c_{a, k}^u$. Both context results involve generating summaries. The process is the same with agent decoder, where two attention distributions are noted as $att_{u, k}^a$, $att_{a, k}^a$.

Since existing models are poor at extracting important information from other roles, it reflects in incorrect cross-role attentions $att_{u, k}^a$ (agent decoder to user context) and $att_{a, k}^u$ (user decoder to agent context). Meanwhile, the same-role attentions $att_{u, k}^u$ (user decoder to user context) and $att_{a, k}^a$ (agent decoder to agent context) are learned better since most information of role-oriented summaries comes from the given role's utterances. Thus we want to use the same-role attention to guide the cross-role attention. As different roles often discuss the same topic in one dialogue, the accumulated cross attention distribution for user decoder and agent decoder on the same role's utterances should be similar. A new \textbf{Attention Divergence Loss} is added to constrain this attention similarity as:

$$\mathcal{L}_{\mathrm{att-user}} = \mathrm{KL}(\mathrm{Avg}(att_u^a) || \mathrm{Avg}(att_u^u))$$
$$\mathcal{L}_{\mathrm{att-agent}} = \mathrm{KL}(\mathrm{Avg}(att_a^u) || \mathrm{Avg}(att_a^a))$$

By minimizing these two losses, the agent decoder attends to user utterances as the user decoder does, and the user decoder attends to agent utterances as the agent decoder does. Two role decoders interactively learn to focus on the key information of the other role in dialogue utterances.

\subsubsection{Decoder Self-Attention Interaction}

Since the decoder calculates the hidden states that could help predict summaries, the hidden states must contain much important information of summaries. We thus try to exploit the information embedded in other role decoders. Specifically, for user decoder, at time step $t$, we achieve the decoder hidden states $h_t^{\mathrm{user}}$ and use a \textbf{Role Attention} module to weigh the last $t$ hidden states of agent decoder $\{h^{\mathrm{agent}}_1, ..., h^{\mathrm{agent}}_t\}$\footnote{Since two decoders decode simultaneously, at step $t$, the other decoder could only provide the states from step $1$ to $t$.}. The role context information $r_t^{\mathrm{user}}$ is obtained by adding all the agent hidden states with their weights, and it helps generate the probability of next word $\hat{y}_t^{\mathrm{user}}$ for user summary. The calculation formulas are given as:
$$r_t^{\mathrm{user}} = \mathrm{Attn}(h_t^{\mathrm{user}}, h_{1:t}^{\mathrm{agent}})$$
$$p(\hat{y}_t^{\mathrm{user}}) = \mathcal{F}(h_t^{\mathrm{user}}, r_t^{\mathrm{user}}, c^u_{u, k}, c^u_{a, k})$$
The function $\mathcal{F}$ includes an MLP layer to fuse different information and a softmax layer to predict the vocabulary probability distribution. The process is the same with the agent decoder, and two decoders decode interactively. 

\subsubsection{Training and Inference}

In the training phase, we use the teacher-forcing method to jointly train two role decoders and use the Negative Log-Likelihood loss to optimize. The NLL loss for a single sample is formulized as:
\begin{align*}
\centering
\mathcal{L}_{\mathrm{nll}} = -(\alpha \cdot &\sum\limits_{i=1}^{|y^{\mathrm{user}}|}{\log P(y^{\mathrm{user}}_i | y^{\mathrm{user}}_{<i}, y^{\mathrm{agent}}_{<i}, D)} + \\
(1-\alpha) \cdot &\sum\limits_{i=1}^{|y^{\mathrm{agent}}|}{\log P(y^{\mathrm{agent}}_i | y^{\mathrm{agent}}_{<i}, y^{\mathrm{user}}_{<i}, D)})
\end{align*}
$\alpha$ is a hyper-parameter for balancing the weights of different summarization tasks. Besides, we add the attention divergence loss to constrain the attention distribution, and the total loss is calculated as:
$$\mathcal{L} = \mathcal{L}_{\mathrm{nll}} + \beta (\mathcal{L}_{\mathrm{att-user}} + \mathcal{L}_{\mathrm{att-agent}})$$
$\beta$ is a hyper-parameter for balancing the weights of different loss functions.

In the inference phase, we also make some adjustments to beam search for our proposed method. We maintain two beams, one for user summary and one for agent summary. At each step of decoding, the $k^{th}$ sequence of the user summary beam should consider the states in the $k^{th}$ sequence of the agent summary beam for role attention. Once one beam has finished decoding, we keep the beam fixed and search for the other one. The beam search will finish if both beams have finished searching.

\section{Experiments}                                         

\subsection{Datasets}

There are two dialogue summarization datasets with role-oriented summarization tasks. Thus, we evaluate the effectiveness of our proposed method on both datasets. First, we experiment on a Chinese fine-grained customer service summarization dataset named CSDS\footnote{https://github.com/xiaolinAndy/CSDS.} \citep{tba}. It provides separate summaries for the user and the agent, and both may contain multiple topics. The other one is a Chinese medical inquiry summarization dataset MC\footnote{https://github.com/cuhksz-nlp/HET-MC. We use the official crawling script to acquire the dataset and divide some data from the training set as the validation set. Due to the website update, the data may have a slight difference compared with the version in the original paper.} \citep{song2020summarizing}. Each dialogue has a summary of the patient's description and a summary of the doctor's suggestion. We note them as user summary and agent summary as well. Most of the summaries in MC are extractive, and only a few are different from dialogue scripts. Moreover, most dialogues in MC have only one topic. Comparing two datasets, MC is easier to summarize while CSDS is more specific for role-oriented summarization and more challenging. The detailed statistics of the two datasets are given in Table \ref{tab:dataset}.

\begin{table}[tbp]
\centering
\small
\begin{tabular}{l|ll}
\hline
                  & \textbf{CSDS} & \textbf{MC} \\ \hline
Train Size        & 9,101         & 29,324      \\
Val Size          & 800           & 3,258       \\
Test Size         & 800           & 8,146       \\
Turns             & 25.92         & 18.48       \\
Dial. Length      & 321.92        & 292.21      \\
User Sum. Length  & 37.28         & 22.37       \\
Agent Sum. Length & 48.08         & 95.32       \\ \hline
\end{tabular}
\caption{\label{tab:dataset} Statistics of CSDS and MC. All the lengths are counted on Chinese characters.}
\end{table}

\subsection{Baselines and Experiment Settings}

We apply the role interaction methods on two widely-used seq2seq models in the summarization community, including PGN \citep{see2017get} and BERTAbs \citep{liu2019text}. Therefore, we will introduce these two backbone models and how we apply \textbf{Role Interactions} to them.

\subsubsection{PGN-based Methods}

PGN is an LSTM-based seq2seq model with a copy mechanism to copy words from the input and a coverage mechanism for constraining context attention. We set two PGN-based baselines for comparison. \textbf{PGN-single} is to separately train two PGN models for generating user summary and agent summary, while \textbf{PGN-multi} tries to jointly train two PGN models by sharing the same encoder. Both baselines adopt all the dialogue context as input.

To apply role interactions, we choose the output of the LSTM cell in the decoder as the query to calculate cross attention and role attention. The attention object in role attention is the output of the LSTM cell from the other decoder. Since we calculate the cross attention for different roles separately, we use a learnable gate $p_{role}$ to control the weight of different cross attentions and add them together according to their weights to achieve the overall context attention distribution. It is also used for the copy and coverage mechanism. We set \textbf{PGN-cross} as adding cross attention interaction, \textbf{PGN-self} as adding decoder self-attention interaction, and \textbf{PGN-both} as adding both interactions.

\subsubsection{BERTAbs-based Methods}

Transformer has been widely used in language understanding and generation models due to its strong representation ability and concurrency, especially in pretrained models \citep{devlin-etal-2019-bert, lewis-etal-2020-bart}. Here we choose BERTAbs \citep{liu2019text} as the backbone structure since it performs well on many summarization datasets and is available for non-English languages such as Chinese. It adopts a pretrained BERT model as encoder and a transformer decoder structure to decode summaries. Both the encoder and the decoder contain six layers, and each layer contains three sub-layers (self-attention, encoder-decoder attention, feedforward). Similar with PGN-based methods, we set \textbf{BERT-single} and \textbf{BERT-multi} as two baselines.

We apply both interactions to each layer in BERTAbs. For cross attention interaction, we change the encoder-decoder attention sub-layer into two separate cross attention modules; for decoder self-attention interaction, we add the role attention module parallel with the encoder-decoder attention module. The query, key, and value of the role attention module are all the output from the self-attention sub-layer. \textbf{BERT-cross}, \textbf{BERT-self}, and \textbf{BERT-both} are kept the same with the settings in PGN-based methods.

\subsubsection{Other Experiment Settings}

We add the role information to the front of the utterance in each turn and concatenate all the utterances in the dialogue sequentially as the input of the model. Both PGN\footnote{https://github.com/atulkum/pointer\_summarizer} and BERTAbs\footnote{https://github.com/nlpyang/PreSumm} baseline methods are adopted from publicly available codes. For PGN-based methods, we use pretrained Chinese word vectors provided by Tencent\footnote{https://ai.tencent.com/ailab/nlp/en/embedding.html}, and the vocabulary size is 10,000. While for BERTAbs-based methods, we use the base version of Chinese BERT-wwm\footnote{https://github.com/ymcui/Chinese-BERT-wwm}. The best checkpoint is chosen based on validation set loss, and we use beam search to decode summaries on the best checkpoint with beam size 5. For choosing hyper-parameters,  since the agent summary is more complex than the user summary in MC, we set $\alpha$ to be 0.2 to give the agent summary more weight. It is set to be 0.5 for CSDS. $\beta$ is set to be 0.5 for PGN and 0.25 for BERTAbs. The hyper-parameter settings are chosen by experimenting on the validation set. More details are given in Appendix A.

\subsection{Evaluation Metrics}

We adopt six common automatic evaluation metrics to evaluate the summary quality. The metrics include traditional n-gram overlapping metrics, such as \textbf{ROUGE} \citep{lin2002manual}, \textbf{BLEU} \citep{papineni2002bleu}, and distributed representation matching metrics, including \textbf{BERTScore} \citep{zhang2019bertscore} and \textbf{MoverScore} \citep{zhao2019moverscore}. We use files2rouge toolkit to calculate the F1 score of ROUGE-1, ROUGE-2, ROUGE-L. More details of evaluation scripts are given in Appendix A.

In addition to automatic metrics, we also compare the summary quality at a fine-grained level through human evaluation. Following the human evaluation process in \citet{tba}, we recruit several volunteers and let them evaluate the summaries in the following aspects: (1) \textbf{Informativeness}: Does the generated summary correctly cover the information in the ground truth summary? (2) \textbf{Non-redundancy}: Does the generated summary not contain repeated, meaningless or unnecessary information? (3) \textbf{Fluency}: Is the generated summary well-formed, semantically complete, and easy to understand? All three aspects are evaluated at the sub-summary level\footnote{We split summaries into different topic segments, and each segment is a sub-summary, same as the process in\citet{tba}.} on a three-point scale, 0 for the worst, 1 for the medium, and 2 for the best.

\begin{table*}[tbp]
\small
\centering
\resizebox{\textwidth}{!}{
\begin{tabular}{l|ll|ll|ll|ll|ll|ll}
\hline
\multicolumn{1}{c|}{\multirow{2}{*}{\textbf{CSDS}}} & \multicolumn{2}{c|}{\textbf{ROUGE-1}}    & \multicolumn{2}{c|}{\textbf{ROUGE-2}}     & \multicolumn{2}{c|}{\textbf{ROUGE-L}}     & \multicolumn{2}{c|}{\textbf{BLEU}}        & \multicolumn{2}{c|}{\textbf{BERTScore}}   & \multicolumn{2}{c}{\textbf{MoverScore}}  \\ \cline{2-13} 
\multicolumn{1}{c|}{}     & \multicolumn{1}{l}{user}           & \multicolumn{1}{l|}{agent}          & \multicolumn{1}{l}{user}           & \multicolumn{1}{l|}{agent}          & \multicolumn{1}{l}{user}           & \multicolumn{1}{l|}{agent}          & \multicolumn{1}{l}{user}           & \multicolumn{1}{l|}{agent}          & \multicolumn{1}{l}{user}           & \multicolumn{1}{l|}{agent}          & \multicolumn{1}{l}{user}           & \multicolumn{1}{l}{agent}          \\ \hline
PGN-single                & 53.55          & 50.20          & 37.06          & 35.12          & 51.05          & 47.59          & 29.64          & 28.25          & 78.68          & 76.13          & 26.68          & 25.13          \\
PGN-multi                 & 54.01          & 49.94          & 37.38          & 34.78          & 51.95          & 48.20          & 30.04          & 29.09          & 78.78          & 75.95          & 27.16          & 24.90          \\ \hline
PGN-cross          & 54.34          & 50.80          & 37.75          & 35.89          & 51.95          & 48.20           & 31.19          & 30.40          & 78.97          & 76.51          & 27.89          & 25.60          \\
PGN-self          & 55.49          & 51.00          & 38.75          & 35.70          & 53.08          & 48.52          & 31.84          & \textbf{30.47} & 79.37          & 76.48          & 27.74          & 25.55          \\
PGN-both      & \textbf{56.08}* & \textbf{51.62}* & \textbf{39.10}* & \textbf{36.50}*  & \textbf{53.89}* & \textbf{49.12}* & \textbf{33.54}* & 29.78*          & \textbf{79.52}* & \textbf{76.74}* & \textbf{28.28}* & \textbf{26.25}* \\ \hline
BERT-single            & 52.72          & 49.57          & 36.39          & 33.82          & 50.44          & 46.83          & 30.17          & 26.99          & 79.23          & 76.39          & 24.96          & 23.87          \\
BERT-multi             & 56.09          & 50.49          & 39.91          & 35.17          & 54.02          & 48.08          & 26.91          & 25.39          & 80.50          & 76.65          & 27.19          & 23.71          \\ \hline
BERT-cross      & 57.29          & 50.35          & \textbf{41.03} & 35.27          & \textbf{55.29} & 48.09          & 30.70          & 24.19          & \textbf{80.90} & 76.65          & 28.55          & 23.70          \\
BERT-self      & 56.94          & 50.96          & 40.37          & 35.24          & 54.85          & 48.40          & 30.61          & 27.13          & 80.53          & 76.80          & 28.24          & 24.83          \\
BERT-both  & \textbf{57.36}* & \textbf{51.92}* & 40.70          & \textbf{36.37}* & 55.17*          & \textbf{49.52}* & \textbf{32.04}* & \textbf{29.23}* & 80.70          & \textbf{77.23}* & \textbf{28.66}* & \textbf{25.48}* \\ \hline
\end{tabular}}
\caption{\label{tab:csds_auto_res} The automatic metric results for CSDS. * indicates that the improvement of applying two interactions (PGN-both, BERT-both) over \emph{single} and \emph{multi} are both statistically significant (p $<$ 0.01).}
\end{table*}

\begin{table*}[tbp]
\small
\centering
\resizebox{\textwidth}{!}{
\begin{tabular}{l|ll|ll|ll|ll|ll|ll}
\hline
\multicolumn{1}{c|}{\multirow{2}{*}{\textbf{MC}}} & \multicolumn{2}{c|}{\textbf{ROUGE-1}}    & \multicolumn{2}{c|}{\textbf{ROUGE-2}}     & \multicolumn{2}{c|}{\textbf{ROUGE-L}}     & \multicolumn{2}{c|}{\textbf{BLEU}}        & \multicolumn{2}{c|}{\textbf{BERTScore}}   & \multicolumn{2}{c}{\textbf{MoverScore}}  \\ \cline{2-13} 
\multicolumn{1}{c|}{}     & \multicolumn{1}{l}{user}           & \multicolumn{1}{l|}{agent}          & \multicolumn{1}{l}{user}           & \multicolumn{1}{l|}{agent}          & \multicolumn{1}{l}{user}           & \multicolumn{1}{l|}{agent}          & \multicolumn{1}{l}{user}           & \multicolumn{1}{l|}{agent}          & \multicolumn{1}{l}{user}           & \multicolumn{1}{l|}{agent}          & \multicolumn{1}{l}{user}           & \multicolumn{1}{l}{agent}          \\ \hline
\citep{song2020summarizing}                      & 92.80                                              & 83.31                                              & 88.97                                              & 75.48                                              & 92.80                                              & 83.29               &-   &  -                             &       -                                            &            -                                    &         -                                        &                -                 \\ \hline
PGN-single                                        & 94.83                    & 82.63                     & 94.32                    & 77.83                     & 94.78                    & 81.51         & 87.66 & 68.10            & 97.60                    & 91.74                     & 90.28                    & 67.95                     \\
PGN-multi                                         & 94.58                    & 83.16                     & 93.98                    & 78.33                     & 94.53                    & 81.96     & 87.23  & \textbf{69.96}                & 97.49                    & 91.92                     & 89.87                    & 68.42                     \\ \hline
PGN-cross    & \textbf{95.12}     & 83.40     & \textbf{94.63}     & 78.60     & \textbf{95.07}     & 82.18     & \textbf{87.99}     & 69.61     & \textbf{97.75}     & 92.07     & \textbf{90.73}     & 69.06 \\
PGN-self     & 95.08   & 83.17     & 94.59     & 78.48     & 95.04     & 82.00     & 87.90     & 69.29     & 97.70     & 91.99     & 90.64     & 68.54 \\
PGN-both    & 95.11*     & \textbf{83.48}*     & 94.59*     & \textbf{78.73}*     & 95.06*     & \textbf{82.28}*     & 87.82*     & 69.63     & 97.71*     & \textbf{92.15}*     & 90.66*     & \textbf{69.24}* \\ \hline
BERT-single  & 95.13   & 81.66    & 94.50     & 76.73     & 95.08     & 80.42     & 87.20     & 64.09     & 97.86     & 91.71     & 90.31     & 68.29 \\
BERT-multi   & 95.18     & 81.20     & 94.61     & 76.37     & 95.13     & 79.97     & 87.38     & 64.83     & 97.90     & 91.51     & 90.71     & 67.55 \\ \hline
BERT-cross     & 95.18     & 81.75     & 94.61     & 77.04     & 95.13     & 80.55     & 87.40     & \textbf{65.63}     & 97.89     & 91.70     & 90.67     & 68.28 \\
BERT-self     & 95.18   & 81.61     & 94.61     & 77.01     & 95.13     & 80.49     & 87.37     & 65.01     & 97.89     & 91.72     & 90.69     & 68.37 \\
BERT-both    & \textbf{95.19}     & \textbf{82.11}*     & \textbf{94.63}     & \textbf{77.49}*     & \textbf{95.14}     & \textbf{80.92}*     & \textbf{87.40}     & 65.40*     & \textbf{97.90}     & \textbf{91.91}*     & \textbf{90.72}     & \textbf{68.95}* \\ \hline

\end{tabular}}
\caption{\label{tab:mc_auto_res} The automatic metric results for MC. * represents the same with the one in Table \ref{tab:csds_auto_res}.}
\end{table*}

\section{Results and Analysis}

\subsection{Automatic Evaluation Results}

First, we present the results of automatic metrics with Student's t-test as significance test in Table \ref{tab:csds_auto_res} and \ref{tab:mc_auto_res}. The results are similar on the two datasets. First, the multi-task mechanism could bring some improvement than separately training on most of the metrics. However, the improvement is limited, especially for the PGN model on CSDS. After adding the enhancement of other roles' information, the performance is significantly boosted. 

On CSDS, PGN-single and BERT-single are two strong baselines provided in \citet{tba}\footnote{Note that we do not mention the baseline Fast-RL \citep{chen2018fast} in \citet{tba}. It first extracts salient utterances and then generates summary sentences from each utterance separately, which is not available to add our proposed interaction methods.}. For PGN-based methods, the best method PGN-both utilizes two interactions and achieves 2.84 and 1.53 higher points on ROUGE-L for user summary and agent summary than PGN-single. For BERTAbs-based methods, the improvements are even greater, which are 4.73 and 2.69. We also conduct ablation studies by only applying one interaction (\emph{-cross} or \emph{-self}). Both settings show promising improvement over the \emph{single} and \emph{multi} baselines on nearly all the metrics, demonstrating the effectiveness of each interaction method. In comparison, applying two interactions together yields the best result on the majority of metrics.

The circumstance is similar on MC. User summarization is relatively simple on MC, and the baseline methods could achieve high performance (5.35 points of ROUGE-2 higher than the best performance in the original paper \citep{song2020summarizing}). Despite this, both cross attention interaction and decoder self-attention interaction could still increase the performance of user summary a bit. Additionally, the improvement on agent summary is more significant. PGN-both method achieves 0.90 points of ROUGE-2 and 1.29 points of MoverScore improvement, while BERT-both achieves 0.76 points of ROUGE-2 and 0.66 points of MoverScore improvement. PGN-both also beats the best result in the original paper on most of the metrics, which uses additional information such as hospital department and disease name. In conclusion, our proposed two interaction methods could bring remarkable improvement on different backbone structures and different datasets.

\begin{table}[]
\small
\centering
\resizebox{\columnwidth}{!}{
\begin{tabular}{l|l|l|l|l}
\hline
\multicolumn{1}{c|}{\textbf{CSDS}} & \multicolumn{1}{c|}{\textbf{Info}} & \multicolumn{1}{c|}{\textbf{Non-Red}} & \multicolumn{1}{c|}{\textbf{Flu}} & \multicolumn{1}{c}{\textbf{Overall}} \\ \hline
PGN-multi                          & \textbf{0.69}/0.65                         & 0.54/0.55                            & 0.70/0.79                        & 0.64/0.66                            \\
PGN-both                           & 0.66/\textbf{0.69}                         & \textbf{0.58}/\textbf{0.59*}                            & \textbf{0.73}/\textbf{0.81}                        & \textbf{0.66}/\textbf{0.70*}                            \\ \hline
BERT-multi                         & 0.58/0.56                         & \textbf{0.66}/\textbf{0.61}                            & 0.84/\textbf{0.87}                        & 0.69/0.68                            \\
BERT-both                          & \textbf{0.62*}/\textbf{0.60*}                         & 0.62/0.60                            & \textbf{0.85}/\textbf{0.87}                        & \textbf{0.70}/\textbf{0.69}                            \\ \hline
\end{tabular}}
\caption{\label{tab:csds_human_res} The human evaluation results for CSDS. Two values in each block represent user summary and agent summary. All the values are in range 0 to 1. * indicates that the improvement of applying two interactions over the \textit{multi} baseline is statistically significant (p < 0.05).}
\end{table}

\begin{table*}[]
\small
\centering
\resizebox{\textwidth}{!}{
\begin{tabular}{l|l|l|l|l|l|l}
\hline
\multicolumn{1}{c|}{\multirow{2}{*}{\textbf{CSDS}}} & \multicolumn{1}{c|}{\textbf{ROUGE-1}} & \multicolumn{1}{c|}{\textbf{ROUGE-2}} & \multicolumn{1}{c|}{\textbf{ROUGE-L}} & \multicolumn{1}{c|}{\textbf{BLEU}} & \multicolumn{1}{c|}{\textbf{BERTScore}} & \multicolumn{1}{c}{\textbf{MoverScore}} \\ 
\multicolumn{1}{c|}{}                               & \multicolumn{1}{c|}{Type A/B}                             & \multicolumn{1}{c|}{Type A/B}                             & \multicolumn{1}{c|}{Type A/B}                             & \multicolumn{1}{c|}{Type A/B}                          & \multicolumn{1}{c|}{Type A/B}                               & \multicolumn{1}{c}{Type A/B}                                \\ \hline
PGN-multi                                          & 55.13/59.45                          & 37.76/41.22                          & 52.73/56.20                          & 30.66/28.29                       & 76.40/77.64                            & 23.74/25.47                             \\
PGN-both                                           & \textbf{56.00/62.28}                 & \textbf{38.58/43.88}                 & \textbf{53.66/58.99}                 & \textbf{31.06/29.14}              & \textbf{76.84/78.59}                   & \textbf{24.41/27.15}                    \\ \hline
BERT-multi                                         & 46.59/50.07                          & 32.33/34.59                          & 44.49/47.65                          & 23.45/26.37                       & 75.03/75.64                            & 22.47/24.34                             \\
BERT-both                                          & \textbf{50.96/54.62}                 & \textbf{35.72/37.93}                 & \textbf{48.82/51.93}                 & \textbf{27.47/30.10}              & \textbf{76.27/76.94}                   & \textbf{24.19/26.17}                    \\ \hline
\end{tabular}}
\caption{\label{tab:complete} The performance on different types of samples. Type A represents agent summaries that need to be integrated, and Type B represents for those that do not. Here all the metrics here are recall scores except for BLEU and MoverScore since they do not have a recall version. We use their available results instead.}
\end{table*}

\subsection{Human Evaluation Results}
To evaluate the summary quality at a more fine-grained level, we compare the summaries from different models according to the pre-defined three aspects: informativeness, non-redundancy, fluency. Since the multi-task framework works better than the single baseline, we directly compare it with applying both interactions. As CSDS is more challenging for this task, we randomly select 100 samples from the test set and obtain the outputs of two baseline methods (PGN-multi and BERT-multi) and two interaction methods (PGN-both and BERT-both). We recruit three volunteers and train them on the evaluation rules\footnote{More details are in Appendix C with ethical concerns.}. Then we let them evaluate the generated summaries according to the ground truth and the original dialogue in the three aspects.  We run the inter-annotator agreement study on three volunteers' scores, and obtain a reasonable kappa score, 0.48 on average. We also calculate an ``Overall'' metric by averaging the results of all three aspects to represent the summary quality in general. We normalize the result into 0 to 1 and present it in Table \ref{tab:csds_human_res}.

The result shows different trends on two backbone structures. For the PGN model, applying interactions could largely reduce the redundancy of both user and agent summary, with a comparable performance of informativeness. Besides, it also improves the fluency of the two summaries. For the BERTAbs model, the interaction method significantly improves the informativeness while the redundancy also increases a bit. The difference exists because BERTAbs prefers to generate short summaries. Thus, considering information from other roles could help generate more useful information but also induce some redundant text. In contrast, PGN tends to generate lengthy summaries. When considering information from other roles, it first tries to discard the redundant texts and only retains more important ones. The fluency improvement on both methods proves that other roles' information helps generate more semantically complete summaries. Considering the overall metric, we conclude that our proposed interaction method is also effective through human evaluation.


\subsection{Further Analysis}
\paragraph{Agent Summary Completeness Analysis}
The agent summary often suffers semantic incompleteness due to missing key information from other roles \citep{tba}. Since our proposed role interactions aim at extracting other roles' key information, we wonder whether they work on these incomplete cases. Following the settings in \citet{tba}, we compare the summary quality of samples that need to integrate other roles' information and those that do not need separately\footnote{It is judged by considering whether the summary needs to refer to other roles’ utterances, which is already labeled in CSDS.}. The result in Table \ref{tab:complete} shows that the interaction method could actually help improve the performance on samples that need to integrate. Besides, samples that do not need also get improved. We believe that it is because considering other roles' information could also help extract critical content from the role's own utterances as well. 


\paragraph{Topic Structural Summary Analysis}
Since we assume that role interactions could help generate better summaries by sharing the same discussion topic, we wonder whether the summaries generated by our methods could include the correct topic structure. More specifically, we want to find out the performance of our methods on summarizing each topic. Following the evaluation method in \cite{tba}, we treat each sentence in the summary as a sub-summary for a single topic and calculate the number of matching sub-summaries with the reference by a ROUGE-L-based matching algorithm. We calculate the precision, recall, and F1 scores of correctly matched sub-summary ratios and present them in Table \ref{tab:topic}. The result shows that two role interaction methods achieve higher recall and F1 scores on sub-summary matching. It proves that role interactions could help the model grasp the discussion topic in the dialogue and generate a more accurate summary for each topic.

We also present an example in Appendix B to prove the effectiveness of our proposed role interaction method.

\begin{table}[]
\small
\centering
\resizebox{\columnwidth}{!}{
\begin{tabular}{l|l|l|l}
\hline
\multicolumn{1}{c|}{\textbf{Methods}} & \multicolumn{1}{c|}{\textbf{Precison}}    & \multicolumn{1}{c|}{\textbf{Recall}}      & \multicolumn{1}{c}{\textbf{F1}}          \\ \hline
PGN-multi        & 28.61/18.86          & 28.87/19.67          & 28.74/19.27          \\
PGN-both         & \textbf{31.79/21.06} & \textbf{30.85/21.58} & \textbf{31.31/21.32} \\ \hline
BERT-multi    & \textbf{40.16/23.99} & 30.26/18.81          & 34.51/21.09          \\
BERT-both     & 37.37/22.09          & \textbf{32.17/20.66} & \textbf{34.57/21.35} \\ \hline
\end{tabular}}
\caption{\label{tab:topic} Sub-summary matching ratio for baselines and our methods. Two values in each block represents user summary and agent summary.}
\end{table}

\section{Conclusion and Future Work}

In this paper, we focus on the role-oriented dialogue summarization task. To fully exploit the information from other roles, we propose two role interaction methods on cross attention and decoder self-attention. The cross attention interaction calculates the context information for different roles separately and uses same-role attention to guide cross-role attention. The decoder self-attention interaction adds a role attention module to attend to other role decoder states interactively. Experiments on two dialogue summarization datasets prove that both interactions perform significantly better than strong baseline methods. Adding role interactions also helps generate summaries with complete semantics and correct topic structure. In the future, we will try to apply this method to other dialogue-related tasks and conduct more experiments on stronger summarization methods.

\section*{Acknowledgements}

We thank all volunteers for their great help on the evaluation, and all anonymous reviewers' suggestions are very appreciated. The research work described in this paper has been partially supported by the National Key Research and Development Program of China under Grant No. 2020AAA0108600 and the Natural Science Foundation of China under Grant No. 62106263.

\bibliography{anthology,custom}

\begin{thebibliography}{33}
\expandafter\ifx\csname natexlab\endcsname\relax\def\natexlab#1{#1}\fi

\bibitem[{Chen and Yang(2020)}]{chen-yang-2020-multi}
Jiaao Chen and Diyi Yang. 2020.
\newblock \href {https://doi.org/10.18653/v1/2020.emnlp-main.336} {Multi-view
  sequence-to-sequence models with conversational structure for abstractive
  dialogue summarization}.
\newblock In \emph{Proceedings of the 2020 Conference on Empirical Methods in
  Natural Language Processing (EMNLP)}, pages 4106--4118, Online. Association
  for Computational Linguistics.

\bibitem[{Chen and Yang(2021)}]{chen-yang-2021-structure}
Jiaao Chen and Diyi Yang. 2021.
\newblock \href {https://doi.org/10.18653/v1/2021.naacl-main.109}
  {Structure-aware abstractive conversation summarization via discourse and
  action graphs}.
\newblock In \emph{Proceedings of the 2021 Conference of the North American
  Chapter of the Association for Computational Linguistics: Human Language
  Technologies}, pages 1380--1391, Online. Association for Computational
  Linguistics.

\bibitem[{Chen and Bansal(2018)}]{chen2018fast}
Yen-Chun Chen and Mohit Bansal. 2018.
\newblock \href {https://doi.org/10.18653/v1/P18-1063} {Fast abstractive
  summarization with reinforce-selected sentence rewriting}.
\newblock In \emph{Proceedings of the 56th Annual Meeting of the Association
  for Computational Linguistics (Volume 1: Long Papers)}, pages 675--686,
  Melbourne, Australia. Association for Computational Linguistics.

\bibitem[{Chen et~al.(2021)Chen, Liu, Chen, and
  Zhang}]{chen-etal-2021-dialogsum}
Yulong Chen, Yang Liu, Liang Chen, and Yue Zhang. 2021.
\newblock \href {https://doi.org/10.18653/v1/2021.findings-acl.449}
  {{D}ialog{S}um: {A} real-life scenario dialogue summarization dataset}.
\newblock In \emph{Findings of the Association for Computational Linguistics:
  ACL-IJCNLP 2021}, pages 5062--5074, Online. Association for Computational
  Linguistics.

\bibitem[{Devlin et~al.(2019)Devlin, Chang, Lee, and
  Toutanova}]{devlin-etal-2019-bert}
Jacob Devlin, Ming-Wei Chang, Kenton Lee, and Kristina Toutanova. 2019.
\newblock \href {https://doi.org/10.18653/v1/N19-1423} {{BERT}: Pre-training of
  deep bidirectional transformers for language understanding}.
\newblock In \emph{Proceedings of the 2019 Conference of the North {A}merican
  Chapter of the Association for Computational Linguistics: Human Language
  Technologies, Volume 1 (Long and Short Papers)}, pages 4171--4186,
  Minneapolis, Minnesota. Association for Computational Linguistics.

\bibitem[{Duan et~al.(2019)Duan, Zhang, Yuan, Zhou, Liu, Wang, Wang, Zhang,
  Sun, and Wu}]{10.1145/3357384.3357940}
Xinyu Duan, Yating Zhang, Lin Yuan, Xin Zhou, Xiaozhong Liu, Tianyi Wang,
  Ruocheng Wang, Qiong Zhang, Changlong Sun, and Fei Wu. 2019.
\newblock \href {https://doi.org/10.1145/3357384.3357940} {Legal summarization
  for multi-role debate dialogue via controversy focus mining and multi-task
  learning}.
\newblock In \emph{Proceedings of the 28th ACM International Conference on
  Information and Knowledge Management}, CIKM '19, page 1361–1370, New York,
  NY, USA. Association for Computing Machinery.

\bibitem[{Feng et~al.(2021)Feng, Feng, and Qin}]{feng2021survey}
Xiachong Feng, Xiaocheng Feng, and Bing Qin. 2021.
\newblock A survey on dialogue summarization: Recent advances and new
  frontiers.
\newblock \emph{arXiv preprint arXiv:2107.03175}.

\bibitem[{Gliwa et~al.(2019)Gliwa, Mochol, Biesek, and Wawer}]{gliwa2019samsum}
Bogdan Gliwa, Iwona Mochol, Maciej Biesek, and Aleksander Wawer. 2019.
\newblock \href {https://doi.org/10.18653/v1/D19-5409} {{SAMS}um corpus: A
  human-annotated dialogue dataset for abstractive summarization}.
\newblock In \emph{Proceedings of the 2nd Workshop on New Frontiers in
  Summarization}, pages 70--79, Hong Kong, China. Association for Computational
  Linguistics.

\bibitem[{Janin et~al.(2003)Janin, Baron, Edwards, Ellis, Gelbart, Morgan,
  Peskin, Pfau, Shriberg, Stolcke et~al.}]{janin2003icsi}
Adam Janin, Don Baron, Jane Edwards, Dan Ellis, David Gelbart, Nelson Morgan,
  Barbara Peskin, Thilo Pfau, Elizabeth Shriberg, Andreas Stolcke, et~al. 2003.
\newblock The icsi meeting corpus.
\newblock In \emph{2003 IEEE International Conference on Acoustics, Speech, and
  Signal Processing, 2003. Proceedings.(ICASSP'03).}, volume~1, pages I--I.
  IEEE.

\bibitem[{Krishna et~al.(2021)Krishna, Khosla, Bigham, and
  Lipton}]{krishna-etal-2021-generating}
Kundan Krishna, Sopan Khosla, Jeffrey Bigham, and Zachary~C. Lipton. 2021.
\newblock \href {https://doi.org/10.18653/v1/2021.acl-long.384} {Generating
  {SOAP} notes from doctor-patient conversations using modular summarization
  techniques}.
\newblock In \emph{Proceedings of the 59th Annual Meeting of the Association
  for Computational Linguistics and the 11th International Joint Conference on
  Natural Language Processing (Volume 1: Long Papers)}, pages 4958--4972,
  Online. Association for Computational Linguistics.

\bibitem[{Lei et~al.(2021)Lei, Yan, Zeng, He, Zhang, and Xu}]{9414547}
Yuejie Lei, Yuanmeng Yan, Zhiyuan Zeng, Keqing He, Ximing Zhang, and Weiran Xu.
  2021.
\newblock \href {https://doi.org/10.1109/ICASSP39728.2021.9414547}
  {Hierarchical speaker-aware sequence-to-sequence model for dialogue
  summarization}.
\newblock In \emph{ICASSP 2021 - 2021 IEEE International Conference on
  Acoustics, Speech and Signal Processing (ICASSP)}, pages 7823--7827.

\bibitem[{Lewis et~al.(2020)Lewis, Liu, Goyal, Ghazvininejad, Mohamed, Levy,
  Stoyanov, and Zettlemoyer}]{lewis-etal-2020-bart}
Mike Lewis, Yinhan Liu, Naman Goyal, Marjan Ghazvininejad, Abdelrahman Mohamed,
  Omer Levy, Veselin Stoyanov, and Luke Zettlemoyer. 2020.
\newblock \href {https://doi.org/10.18653/v1/2020.acl-main.703} {{BART}:
  Denoising sequence-to-sequence pre-training for natural language generation,
  translation, and comprehension}.
\newblock In \emph{Proceedings of the 58th Annual Meeting of the Association
  for Computational Linguistics}, pages 7871--7880, Online. Association for
  Computational Linguistics.

\bibitem[{Lin and Hovy(2002)}]{lin2002manual}
Chin-Yew Lin and Eduard Hovy. 2002.
\newblock \href {https://doi.org/10.3115/1118162.1118168} {Manual and automatic
  evaluation of summaries}.
\newblock In \emph{Proceedings of the {ACL}-02 Workshop on Automatic
  Summarization}, pages 45--51, Phildadelphia, Pennsylvania, USA. Association
  for Computational Linguistics.

\bibitem[{Lin et~al.(2021)Lin, Ma, Zhu, Xiang, Zhou, Zhang, and Zong}]{tba}
Haitao Lin, Liqun Ma, Junnan Zhu, Lu~Xiang, Yu~Zhou, Jiajun Zhang, and
  Chengqing Zong. 2021.
\newblock \href {https://aclanthology.org/2021.emnlp-main.365} {{CSDS}: A
  fine-grained {C}hinese dataset for customer service dialogue summarization}.
\newblock In \emph{Proceedings of the 2021 Conference on Empirical Methods in
  Natural Language Processing}, pages 4436--4451, Online and Punta Cana,
  Dominican Republic. Association for Computational Linguistics.

\bibitem[{Liu et~al.(2019)Liu, Wang, Xu, Li, and Ye}]{liu2019automatic}
Chunyi Liu, Peng Wang, Jiang Xu, Zang Li, and Jieping Ye. 2019.
\newblock Automatic dialogue summary generation for customer service.
\newblock In \emph{Proceedings of the 25th ACM SIGKDD International Conference
  on Knowledge Discovery \& Data Mining}, pages 1957--1965.

\bibitem[{Liu and Lapata(2019)}]{liu2019text}
Yang Liu and Mirella Lapata. 2019.
\newblock \href {https://doi.org/10.18653/v1/D19-1387} {Text summarization with
  pretrained encoders}.
\newblock In \emph{Proceedings of the 2019 Conference on Empirical Methods in
  Natural Language Processing and the 9th International Joint Conference on
  Natural Language Processing (EMNLP-IJCNLP)}, pages 3730--3740, Hong Kong,
  China. Association for Computational Linguistics.

\bibitem[{Liu et~al.(2020)Liu, Zhang, Xiong, Zhou, He, Wu, Wang, and
  Zong}]{Liu_Zhang_Xiong_Zhou_He_Wu_Wang_Zong_2020}
Yuchen Liu, Jiajun Zhang, Hao Xiong, Long Zhou, Zhongjun He, Hua Wu, Haifeng
  Wang, and Chengqing Zong. 2020.
\newblock \href {https://doi.org/10.1609/aaai.v34i05.6360} {Synchronous speech
  recognition and speech-to-text translation with interactive decoding}.
\newblock \emph{Proceedings of the AAAI Conference on Artificial Intelligence},
  34(05):8417--8424.

\bibitem[{Liu et~al.(2021)Liu, Shi, and Chen}]{liu-etal-2021-coreference}
Zhengyuan Liu, Ke~Shi, and Nancy Chen. 2021.
\newblock \href {https://aclanthology.org/2021.sigdial-1.53} {Coreference-aware
  dialogue summarization}.
\newblock In \emph{Proceedings of the 22nd Annual Meeting of the Special
  Interest Group on Discourse and Dialogue}, pages 509--519, Singapore and
  Online. Association for Computational Linguistics.

\bibitem[{McCowan et~al.(2005)McCowan, Carletta, Kraaij, Ashby, Bourban, Flynn,
  Guillemot, Hain, Kadlec, Karaiskos et~al.}]{mccowan2005ami}
I~McCowan, J~Carletta, W~Kraaij, S~Ashby, S~Bourban, M~Flynn, M~Guillemot,
  T~Hain, J~Kadlec, V~Karaiskos, et~al. 2005.
\newblock The ami meeting corpus.
\newblock In \emph{Proceedings of Measuring Behavior 2005, 5th International
  Conference on Methods and Techniques in Behavioral Research}, pages 137--140.
  Noldus Information Technology.

\bibitem[{Mehdad et~al.(2014)Mehdad, Carenini, and
  Ng}]{mehdad-etal-2014-abstractive}
Yashar Mehdad, Giuseppe Carenini, and Raymond~T. Ng. 2014.
\newblock \href {https://doi.org/10.3115/v1/P14-1115} {Abstractive
  summarization of spoken and written conversations based on phrasal queries}.
\newblock In \emph{Proceedings of the 52nd Annual Meeting of the Association
  for Computational Linguistics (Volume 1: Long Papers)}, pages 1220--1230,
  Baltimore, Maryland. Association for Computational Linguistics.

\bibitem[{Papineni et~al.(2002)Papineni, Roukos, Ward, and
  Zhu}]{papineni2002bleu}
Kishore Papineni, Salim Roukos, Todd Ward, and Wei-Jing Zhu. 2002.
\newblock \href {https://doi.org/10.3115/1073083.1073135} {{B}leu: a method for
  automatic evaluation of machine translation}.
\newblock In \emph{Proceedings of the 40th Annual Meeting of the Association
  for Computational Linguistics}, pages 311--318, Philadelphia, Pennsylvania,
  USA. Association for Computational Linguistics.

\bibitem[{See et~al.(2017)See, Liu, and Manning}]{see2017get}
Abigail See, Peter~J. Liu, and Christopher~D. Manning. 2017.
\newblock \href {https://doi.org/10.18653/v1/P17-1099} {Get to the point:
  Summarization with pointer-generator networks}.
\newblock In \emph{Proceedings of the 55th Annual Meeting of the Association
  for Computational Linguistics (Volume 1: Long Papers)}, pages 1073--1083,
  Vancouver, Canada. Association for Computational Linguistics.

\bibitem[{Song et~al.(2020)Song, Tian, Wang, and Xia}]{song2020summarizing}
Yan Song, Yuanhe Tian, Nan Wang, and Fei Xia. 2020.
\newblock \href {https://doi.org/10.18653/v1/2020.coling-main.63} {Summarizing
  medical conversations via identifying important utterances}.
\newblock In \emph{Proceedings of the 28th International Conference on
  Computational Linguistics}, pages 717--729, Barcelona, Spain (Online).
  International Committee on Computational Linguistics.

\bibitem[{Tamura et~al.(2011)Tamura, Ishikawa, Saikou, and
  Tsuchida}]{tamura-etal-2011-extractive}
Akihiro Tamura, Kai Ishikawa, Masahiro Saikou, and Masaaki Tsuchida. 2011.
\newblock \href {https://aclanthology.org/I11-1056} {Extractive summarization
  method for contact center dialogues based on call logs}.
\newblock In \emph{Proceedings of 5th International Joint Conference on Natural
  Language Processing}, pages 500--508, Chiang Mai, Thailand. Asian Federation
  of Natural Language Processing.

\bibitem[{Wang and Cardie(2013)}]{wang-cardie-2013-domain}
Lu~Wang and Claire Cardie. 2013.
\newblock \href {https://aclanthology.org/P13-1137} {Domain-independent
  abstract generation for focused meeting summarization}.
\newblock In \emph{Proceedings of the 51st Annual Meeting of the Association
  for Computational Linguistics (Volume 1: Long Papers)}, pages 1395--1405,
  Sofia, Bulgaria. Association for Computational Linguistics.

\bibitem[{Wang et~al.(2019)Wang, Zhang, Zhou, Liu, and
  Zong}]{wang-etal-2019-synchronously}
Yining Wang, Jiajun Zhang, Long Zhou, Yuchen Liu, and Chengqing Zong. 2019.
\newblock \href {https://doi.org/10.18653/v1/D19-1330} {Synchronously
  generating two languages with interactive decoding}.
\newblock In \emph{Proceedings of the 2019 Conference on Empirical Methods in
  Natural Language Processing and the 9th International Joint Conference on
  Natural Language Processing (EMNLP-IJCNLP)}, pages 3350--3355, Hong Kong,
  China. Association for Computational Linguistics.

\bibitem[{Zhang et~al.(2020)Zhang, Kishore, Wu, Weinberger, and
  Artzi}]{zhang2019bertscore}
Tianyi Zhang, Varsha Kishore, Felix Wu, Kilian~Q. Weinberger, and Yoav Artzi.
  2020.
\newblock \href {https://openreview.net/forum?id=SkeHuCVFDr} {Bertscore:
  Evaluating text generation with {BERT}}.
\newblock In \emph{8th International Conference on Learning Representations,
  {ICLR} 2020, Addis Ababa, Ethiopia, April 26-30, 2020}. OpenReview.net.

\bibitem[{Zhang et~al.(2021)Zhang, Zhang, Zaheer, and
  Ahmed}]{Zhang_Zhang_Zaheer_Ahmed_2021}
Xinyuan Zhang, Ruiyi Zhang, Manzil Zaheer, and Amr Ahmed. 2021.
\newblock \href {https://ojs.aaai.org/index.php/AAAI/article/view/17703}
  {Unsupervised abstractive dialogue summarization for tete-a-tetes}.
\newblock \emph{Proceedings of the AAAI Conference on Artificial Intelligence},
  35(16):14489--14497.

\bibitem[{Zhao et~al.(2019)Zhao, Peyrard, Liu, Gao, Meyer, and
  Eger}]{zhao2019moverscore}
Wei Zhao, Maxime Peyrard, Fei Liu, Yang Gao, Christian~M. Meyer, and Steffen
  Eger. 2019.
\newblock \href {https://doi.org/10.18653/v1/D19-1053} {{M}over{S}core: Text
  generation evaluating with contextualized embeddings and earth mover
  distance}.
\newblock In \emph{Proceedings of the 2019 Conference on Empirical Methods in
  Natural Language Processing and the 9th International Joint Conference on
  Natural Language Processing (EMNLP-IJCNLP)}, pages 563--578, Hong Kong,
  China. Association for Computational Linguistics.

\bibitem[{Zhong et~al.(2021)Zhong, Yin, Yu, Zaidi, Mutuma, Jha, Awadallah,
  Celikyilmaz, Liu, Qiu, and Radev}]{zhong-etal-2021-qmsum}
Ming Zhong, Da~Yin, Tao Yu, Ahmad Zaidi, Mutethia Mutuma, Rahul Jha,
  Ahmed~Hassan Awadallah, Asli Celikyilmaz, Yang Liu, Xipeng Qiu, and Dragomir
  Radev. 2021.
\newblock \href {https://doi.org/10.18653/v1/2021.naacl-main.472} {{QMS}um: A
  new benchmark for query-based multi-domain meeting summarization}.
\newblock In \emph{Proceedings of the 2021 Conference of the North American
  Chapter of the Association for Computational Linguistics: Human Language
  Technologies}, pages 5905--5921, Online. Association for Computational
  Linguistics.

\bibitem[{Zhou et~al.(2019)Zhou, Zhang, and Zong}]{10.1162/tacl_a_00256}
Long Zhou, Jiajun Zhang, and Chengqing Zong. 2019.
\newblock \href {https://doi.org/10.1162/tacl_a_00256} {{Synchronous
  Bidirectional Neural Machine Translation}}.
\newblock \emph{Transactions of the Association for Computational Linguistics},
  7:91--105.

\bibitem[{Zhu et~al.(2020)Zhu, Xu, Zeng, and
  Huang}]{zhu-etal-2020-hierarchical}
Chenguang Zhu, Ruochen Xu, Michael Zeng, and Xuedong Huang. 2020.
\newblock \href {https://doi.org/10.18653/v1/2020.findings-emnlp.19} {A
  hierarchical network for abstractive meeting summarization with cross-domain
  pretraining}.
\newblock In \emph{Findings of the Association for Computational Linguistics:
  EMNLP 2020}, pages 194--203, Online. Association for Computational
  Linguistics.

\bibitem[{Zou et~al.(2021)Zou, Zhao, Kang, Lin, Peng, Jiang, Sun, Zhang, Huang,
  and Liu}]{Zou_Zhao_Kang_Lin_Peng_Jiang_Sun_Zhang_Huang_Liu_2021}
Yicheng Zou, Lujun Zhao, Yangyang Kang, Jun Lin, Minlong Peng, Zhuoren Jiang,
  Changlong Sun, Qi~Zhang, Xuanjing Huang, and Xiaozhong Liu. 2021.
\newblock \href {https://ojs.aaai.org/index.php/AAAI/article/view/17723}
  {Topic-oriented spoken dialogue summarization for customer service with
  saliency-aware topic modeling}.
\newblock \emph{Proceedings of the AAAI Conference on Artificial Intelligence},
  35(16):14665--14673.

\end{thebibliography}
\bibliographystyle{acl_natbib}

\appendix

\section*{Appendix}

\section{Experiment Details}

Here, we will introduce some detailed settings for our experiments on two datasets.
\begin{itemize}
    \item \textbf{PGN-based methods}: We construct the vocabulary by choosing the top 10,000 most frequent words in the training data. The settings of PGN are the same as the original setting with hidden size 256. The optimizer is Adagrad and the learning rate is 0.15.
    
    For CSDS dataset, we use the given word split result to construct the vocabulary. The maximum input length is set as 500. The maximum output length is 100, and the minimum is 10. We train 40 epochs without coverage mechanism or KL divergence loss (if needed) and 10 epochs with coverage mechanism and KL divergence loss. Then we choose the best checkpoint by comparing the loss on the validation set and use it to decode summaries.
    
    For MC dataset, we use jieba\footnote{https://pypi.org/project/jieba/} tool to split sentences into words for constructing the vocabulary. The length conditions of input and output are kept the same with CSDS. We train 30 epochs without coverage mechanism and do not finetune with coverage mechanism since we found that it makes the performance worse. The KL divergence loss is added to the training loss for PGN-cross and PGN-both in all the training process.
    
    \item \textbf{BERTAbs-based methods}: Since the BERT model is already finetuned, there is no need to reconstruct the vocabulary. The Chinese BERT model works on character-level. Thus we set the length limit larger. The dimension and optimizer settings of BERTAbs are also the same as the original settings.  
    
    For CSDS dataset, the maximum input and output length are 1,000 and 200, respectively. The minimum output length is 15. We train the model for 4000 steps and save the checkpoint for every 400 steps. We use Adam optimizer with a warmup of 1000 steps. The KL divergence loss is added by finetuning 1000 more steps. During the inference time, we control the maximum non-repeat n-gram length as 5.
    
    For MC dataset, the maximum input and output length are kept the same as in CSDS, and the minimum output length is 10. We train the model for 8000 steps and add the KL divergence loss in all the training process.
\end{itemize}

All the PGN-based models are run on an NVIDIA TITAN Xp, and all the BERTAbs-based models are run on an NVIDIA RTX3090. The whole running time is less than a week.

We also provide the running scripts of auto evaluation metrics for better reproduction. For ROUGE metrics, we use the files2rouge\footnote{https://github.com/pltrdy/files2rouge} toolkit with the default parameters. All the Chinese characters are transferred into number ids for calculation, and the period is used to split each summary into several sentences for ROUGE-L calculation. For BERTScore, we use the official code\footnote{https://pypi.org/project/bert-score/0.2.1/}. For MoverScore, we use moverscore-v2\footnote{https://github.com/AIPHES/emnlp19-moverscore} and the bert-base-chinese pretrained model for obtaining representations.

\section{Case Study}

\begin{figure*}[tbp]
\centering
\includegraphics[width=0.9\textwidth]{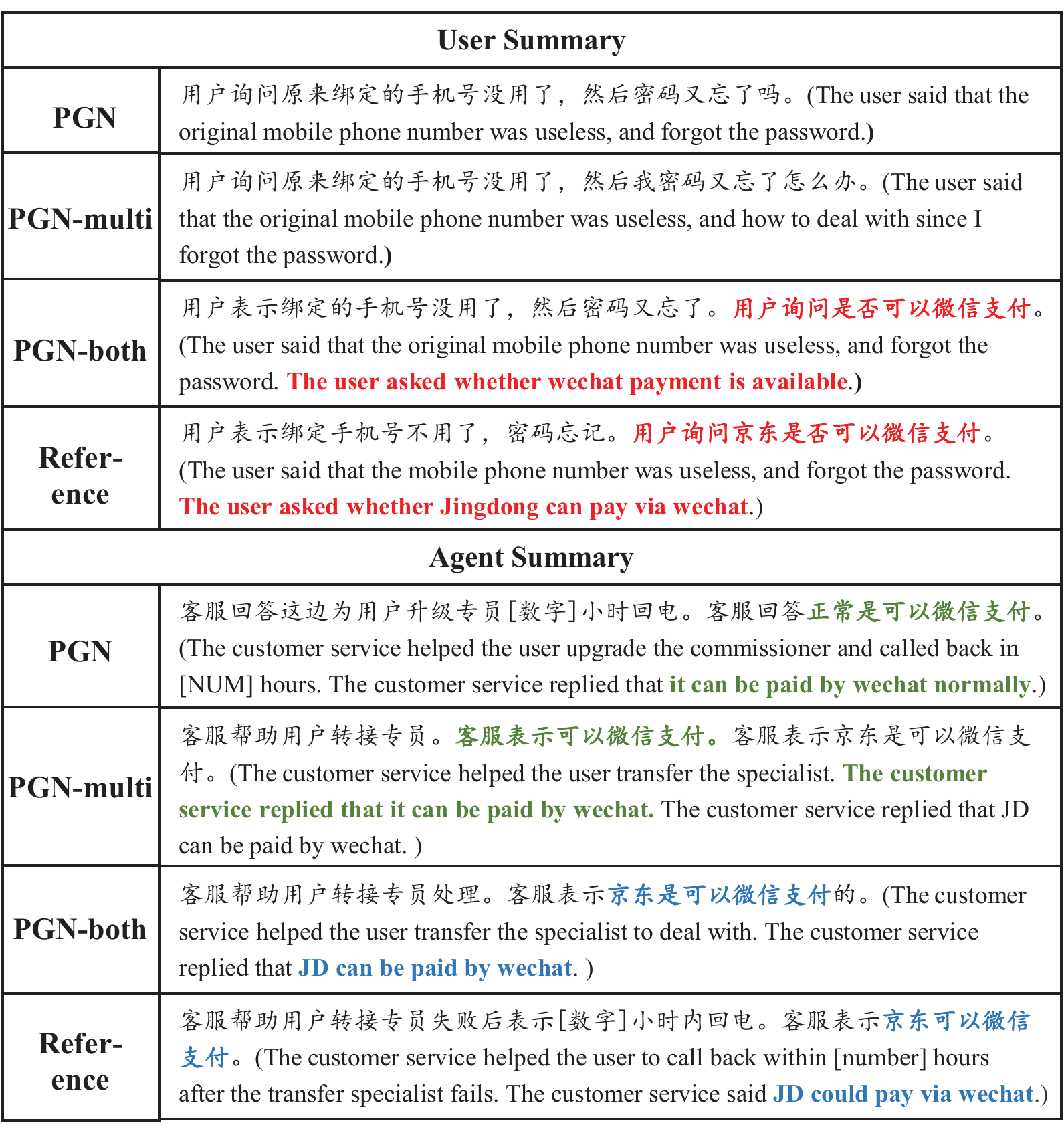} 
\caption{The generated summaries and ground truth for the example dialogue.}
\label{fig:case_sum}
\end{figure*}

Here we use the same example illustrated in the main paper to prove the effectiveness of our proposed method. The outputs of different methods are given in Figure \ref{fig:case_sum}. Comparing the outputs of user summary, only PGN-both correctly summarizes the key issue ``\textit{The user asked whether wechat payment is available.}'', while other baselines omit it. This could be contributed to decoder self-attention interaction by attending to the information in the decoded agent summary. As for the agent summary, PGN omits the keyword ``\textit{JD}'' and PGN-multi generates a redundant clause ``\textit{The customer service replied that it can be paid by wechat}''. Compared with them, PGN-both generates the agent summary with both informativeness and preciseness.

\begin{figure*}
\centering
\subfigure[The average attention distribution for user summary]{
\begin{minipage}[b]{0.9\textwidth}
\includegraphics[width=0.9\textwidth]{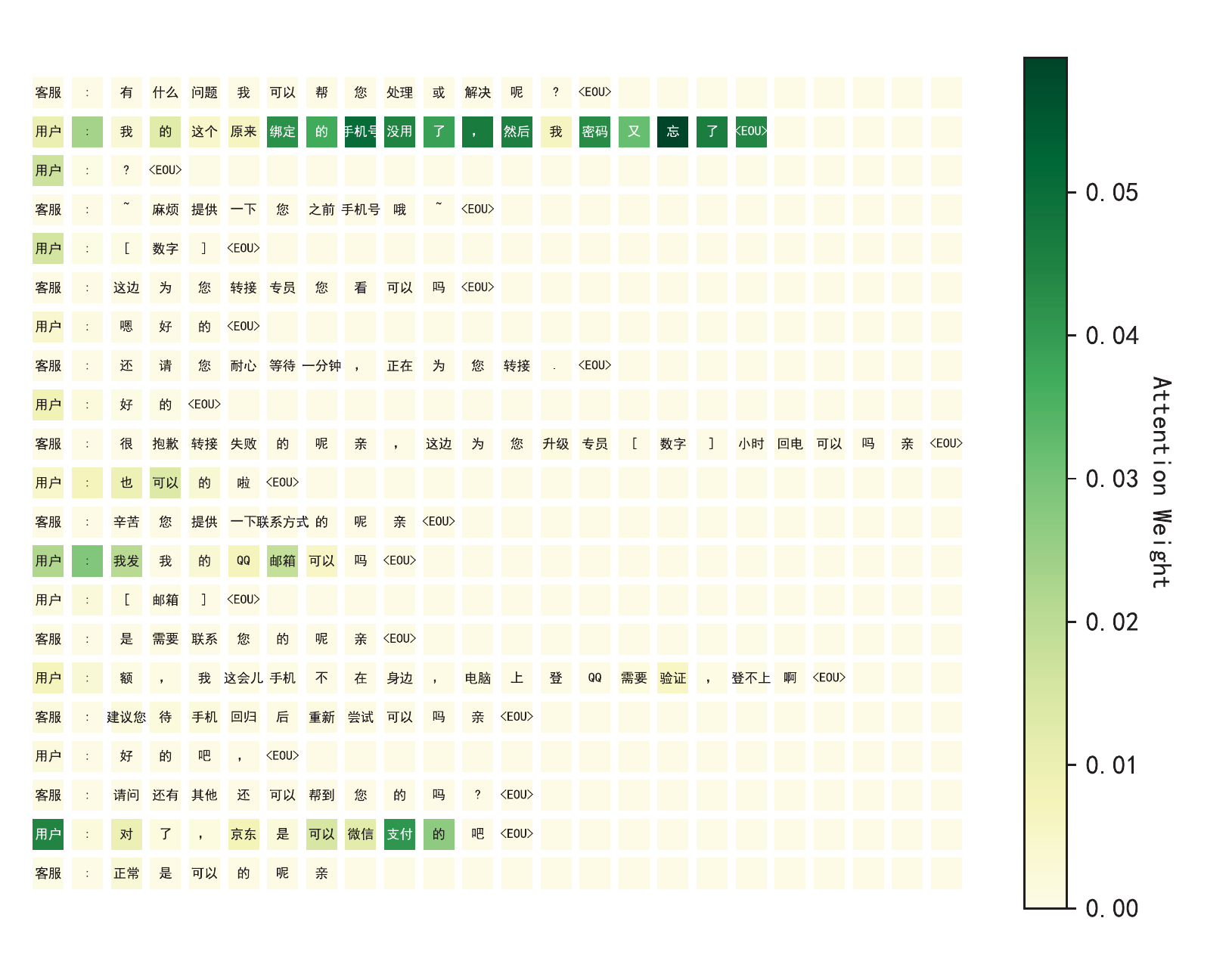}
\end{minipage}
}
\subfigure[The average attention distribution for agent summary]{
\begin{minipage}[b]{0.9\textwidth}
\includegraphics[width=0.9\textwidth]{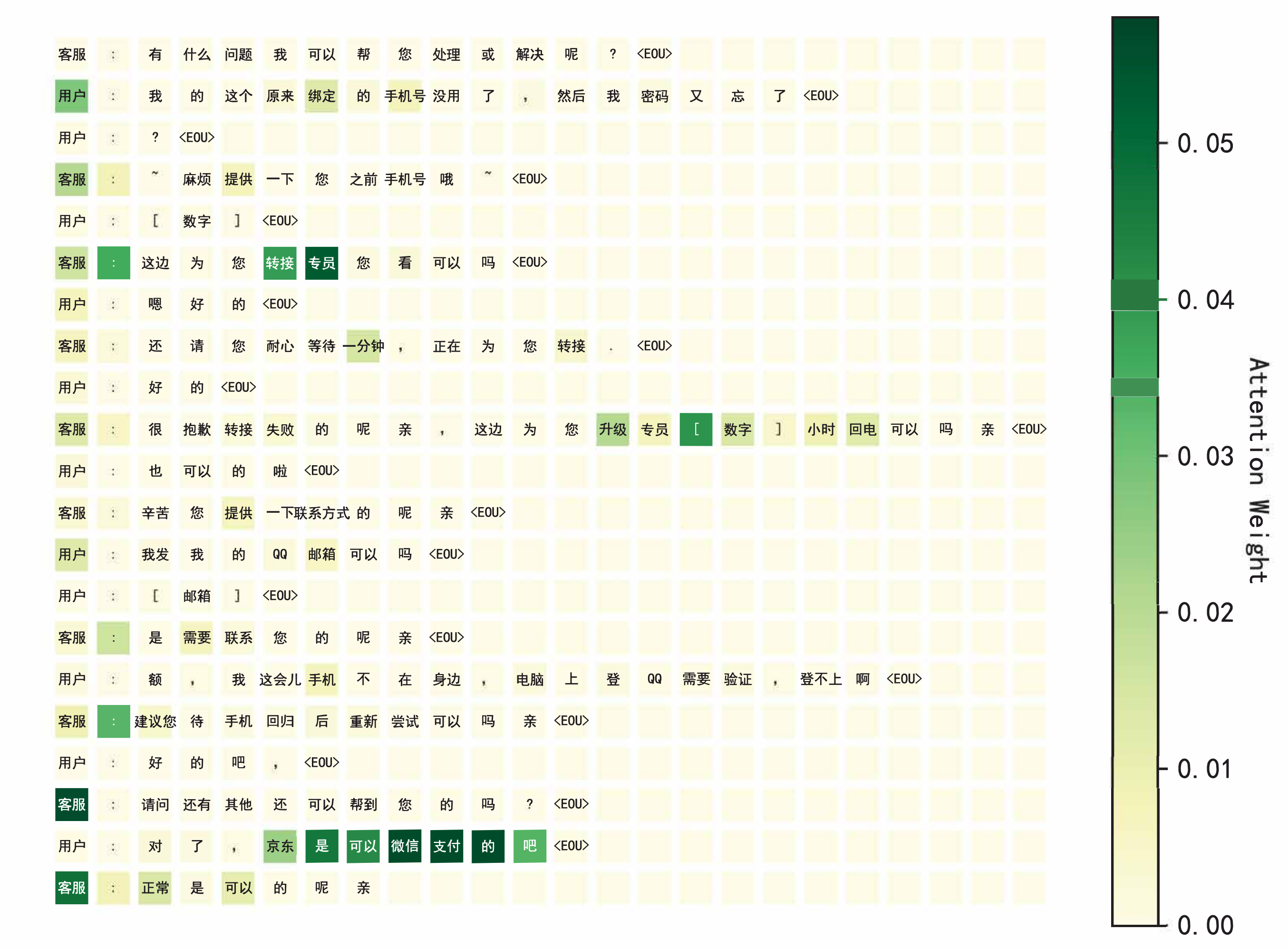}
\end{minipage}
}
\caption{The average attention distribution for decoding summaries using the PGN-both method.} \label{fig:case_dial}
\end{figure*}

As given in Figure \ref{fig:case_dial}, we also present the average attention distribution for cross attention module in the PGN-both method. Although the attention distributions for user summary and agent summary are different, they also show some similarities, such as both focusing on the second line from the bottom, which is the key utterance for both summaries. This could be benefited by the cross attention interaction since we close the gap between the attention distributions of different role decoders on the same role's utterances. 

\section{Ethical Concerns}

We only use the data provided by two datasets for training. The private information in CSDS has already been anonymized, such as replacing all numbers with special token $<$NUM$>$ and all order IDs with $<$ORDER-ID$>$. There is no personal information available in CSDS. The circumstance is the same for MC, where all the dialogues do not contain detailed personal information. Thus the methods provided in our experiment do not have any issues with privacy disclosure. As for human evaluation, all the participants are Chinese graduate students who volunteer to make the evaluation, and they are all proficient in Chinese. We first let them read the evaluating instructions and let them evaluate ten samples without showing the model name for each summary. After confirming that the results provided by three volunteers attain a moderate level of agreement on all the aspects, we allow them to examine the remaining samples.

\end{document}